\documentclass[10pt,twocolumn,letterpaper]{article}
\usepackage[pagenumbers]{cvpr} 

\definecolor{cvprblue}{rgb}{0.21,0.49,0.74}
\usepackage[pagebackref,breaklinks,colorlinks,allcolors=cvprblue]{hyperref}
\def\paperID{6117}
\def\confName{CVPR}
\def\confYear{2025}

\usepackage[accsupp]{axessibility}  

\usepackage{algorithmic}
\usepackage{algorithm}

\newcommand{\alisuregood}[1]{\textcolor{red}{#1}}
\newcommand{\alisurebed}[1]{\textcolor{green}{#1}}
\usepackage{bm}

\usepackage{multirow}
\usepackage{booktabs}
\usepackage{makecell}
\usepackage{hhline}

\usepackage{pifont}
\usepackage{graphicx}

\title{Logits DeConfusion with CLIP for Few-Shot Learning}

\author{Shuo Li, Fang Liu\thanks{Corresponding author}, Zehua Hao, Xinyi Wang, Lingling Li, Xu Liu, Puhua Chen, Wenping Ma\\
Key Laboratory of Intelligent Perception and Image Understanding of Ministry of Education, \\
International Research Center for Intelligent Perception and Computation, \\
Joint International Research Laboratory of Intelligent Perception and Computation, \\ 
School of Artificial Intelligence, Xidian University, Xi'an 710071, China \\
{\tt\small lishuo@xidian.edu.cn, f63liu@163.com}
}

\begin{document}
\maketitle

\begin{abstract}
With its powerful visual-language alignment capability, CLIP performs well in zero-shot and few-shot learning tasks. However, we found in experiments that CLIP's logits suffer from serious inter-class confusion problems in downstream tasks, and the ambiguity between categories seriously affects the accuracy. To address this challenge, we propose a novel method called Logits DeConfusion, which effectively learns and eliminates inter-class confusion in logits by combining our Multi-level Adapter Fusion (MAF) module with our Inter-Class Deconfusion (ICD) module. Our MAF extracts features from different levels and fuses them uniformly to enhance feature representation. Our ICD learnably eliminates inter-class confusion in logits with a residual structure. Experimental results show that our method can significantly improve the classification performance and alleviate the inter-class confusion problem. The code is available at https://github.com/LiShuo1001/LDC.
\end{abstract}

\section{Introduction}
\label{sec:intro}
Recently, the outstanding performance of Vision-Language Models (VLMs) \cite{zhang2024vision} in visual understanding has attracted widespread attention \cite{martin2024transductive}. CLIP (Contrastive Language-Image Pretraining) \cite{radford2021learning}, as a representative model, successfully maps images and texts to a common embedding space through large-scale image-text contrastive learning, and achieves Zero-Shot Learning (ZSL) on unseen categories, thereby reducing the dependence on labeled data and demonstrating excellent transfer learning capabilities \cite{xiao2024florence}. In Few-Shot Learning (FSL), the goal is to quickly adapt to classification tasks of new categories with very few labeled training samples \cite{zhai2022lit}. CLIP shows unique advantages in such tasks, and its pre-trained rich features help improve the generalization ability of the model \cite{lee2022uniclip}. However, although CLIP-based FSL has great potential \cite{zhang2022tip,shao2024deil}, its logits often show serious inter-class confusion, resulting in a decrease in classification accuracy, which limits the performance improvement of applying CLIP to FSL \cite{marchisio2024understanding}.

\begin{figure}
    \centering
    \includegraphics[width=1.0\linewidth]{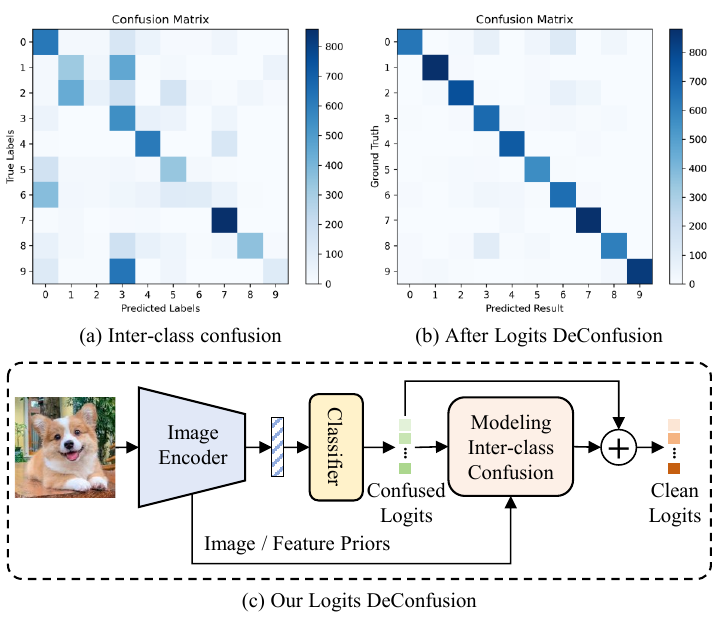}
    \vspace{-5mm}
    \caption{(a) Inter-class confusion of logits in CLIP-based ZSL. (b) After remove inter-class confusion of logits. (c) Our Logits DeConfusion models inter-class confusion and removes it.} 
    \label{fig:intr}
\end{figure}

In the original CLIP-based ZSL \cite{radford2021learning} experiment, as shown in Figure \ref{fig:intr} (a), we found that its logits have significant inter-class confusion, which is manifested as the difficulty in accurately distinguishing the predicted values of different categories in downstream tasks. The main reason for this phenomenon is \textit{the pre-training strategy of CLIP}. CLIP is pre-trained on large-scale image-text pairs via contrast learning, rather than directly optimizing the classification boundaries \cite{radford2021learning}. Therefore, the ability to distinguish categories in classification tasks is insufficient, which leads to obvious inter-class confusion in logits. In addition, \textit{the significant domain differences between the downstream data and the CLIP pre-training data} further exacerbate the confusion between logits of different categories. This inter-class confusion has a serious negative impact on CLIP-based FSL, and it is difficult for the model to learn a reliable classifier from a small number of samples. Especially when the similarity between categories is high, the inter-class confusion problem will be further exacerbated. Therefore, \textit{how to effectively eliminate the inter-class confusion and improve the performance of CLIP-based FSL has become an important challenge that needs to be solved}.

To address the above challenge in CLIP-based FSL \cite{wu2023feature,tang2024amu}, we propose to \textit{model and eliminate inter-class confusion through learning}. Specifically, we first use a small number of samples to model inter-class confusion through a learnable module, which aims to capture inter-class confusion pattern from CLIP's logit using image features as prior information, and then eliminate these confusions through a residual structure to obtain new logits with less confusion. The advantage of modeling and eliminating inter-class confusion is that it can not only \textit{adaptively learn the confusion pattern between classes}, but also \textit{eliminate this confusion through a residual structure}, thereby making the model have stronger generalization ability and accuracy.

In our work, based on the above motivation, as shown in Figure \ref{fig:intr} (c), we propose a method called Logits DeConfusion (LDC), which includes a Multi-level Adapter Fusion (MAF) module and an Inter-Class Deconfusion (ICD) module. First, the MAF module extracts and fuses features from different levels of the CLIP image encoder to construct a unified feature representation, thereby improving adaptability to FSL tasks. Second, the ICD module introduces visual features to provide prior guidance on inter-class confusion, and learns the inter-class confusion patterns through a learnable module, and then removes them through a residual structure to obtain clearer category distinction. As shown in Figure \ref{fig:intr} (b), our method can not only effectively alleviate the inter-class confusion problem and improve classification performance, but also retain the rich feature expression of CLIP and enhance its generalization ability and robustness. In summary, our main contributions are as follows:
\begin{itemize}
    \item A novel Logits DeConfusion method is proposed, which combines a multi-level adapter fusion module and an inter-class deconfusion module to alleviate the inter-class confusion of CLIP-based FSL through learning. 
    \item A multi-level adapter fusion module is designed, which extract and fuse features from different levels of the CLIP image encoder to enhance the image feature representation ability of our method.
    \item Our proposed inter-class deconfusion module effectively models and eliminates inter-class confusion patterns from visual features and CLIP logits through a learnable module with residuals, improving the ability to distinguish categories in downstream FSL tasks.
    \item Experimental results on multiple benchmarks demonstrate that our method significantly improves the classification performance and exhibits stronger generalization. 
\end{itemize}

\section{Related Work}
\label{sec:rel}

In this section, we mainly introduce related work from the following two aspects: Zero/Few-shot Learning (Section \ref{rw_1}) and CLIP-based Zero/Few-shot Learning (Section \ref{rw_2}).

\subsection{Zero/Few-shot Learning}
\label{rw_1}

Zero-Shot Learning (ZSL) and Few-Shot Learning (FSL) are important methods proposed in the field of machine learning to deal with insufficient labeled data \cite{pourpanah2022review}. 
In ZSL, the model needs to classify unseen categories without any labeled data during training \cite{xian2017zero}. 
To address this challenge, early methods often rely on semantic embeddings (such as word embeddings \cite{zhang2017learning} or attribute features) to transfer knowledge from known categories to unseen categories \cite{romera2015embarrassingly}. In addition, Generative Adversarial Networks (GANs) and Variational AutoEncoders (VAEs) have been introduced into the field of ZSL to improve classification capabilities by generating virtual samples of unseen categories \cite{verma2018generalized}. Although these methods have improved generalization capabilities to a certain extent, they often rely on accurate semantic representations and labeling information, which makes it difficult to generalize in complex scenarios \cite{song2023comprehensive}. 
In FSL, models that can quickly adapt to new tasks are designed to solve learning problems with only a small number of labeled samples \cite{DU202213,10820843}. Meta-learning is a classic method in FSL that simulates few-shot scenarios in the training phase so that the model can efficiently adapt to new categories of learning tasks in the testing phase \cite{sun2019meta}. In addition, metric learning methods \cite{li2023knowledge} learn a metric space suitable for few-shot tasks so that samples of new categories are well distinguishable in this space. Although FSL has made progress in tasks such as classification \cite{li2022unsupervised,jiao2024multiscale} and detection \cite{kohler2023few}, it still has the problem of insufficient generalization ability when faced with high similarity between categories \cite{cao2021few} or large domain differences \cite{zhou2023revisiting}. 

\subsection{CLIP-based Zero/Few-shot Learning}
\label{rw_2}

The introduction of CLIP brings breakthrough progress to ZSL and FSL \cite{radford2021learning,jiao2024brain}. 
In ZSL, CLIP can achieve direct classification without labeled data by comparing images with natural language descriptions of categories \cite{zhou2023zegclip,tang2024data}. This property has enabled CLIP to achieve significant performance improvements on multiple datasets and has triggered a large number of follow-up studies on cross-modal representation learning \cite{li2023minent,qian2024intra}. Some subsequent methods such as CoOp (Context Optimization) \cite{zhou2022learning} and CoCoOp (Conditional CoOp) \cite{zhou2022conditional} have further improved the adaptability of CLIP in different tasks by introducing learnable prompts \cite{jiao2019survey}. However, although CLIP performs well in ZSL, it is easy to cause confusion between categories in downstream tasks because it does not directly optimize the classification boundaries during the pre-training process \cite{9810916,wang2023improving}. 
Recently, CLIP-based FSL has received widespread attention \cite{song2022clip,tang2024amu}. Studies have shown that CLIP can improve classification performance in FSL tasks through its rich visual representation and cross-modal learning capabilities \cite{li2022learning,palanisamy2023proto,shao2024deil}. Many methods adaptively adjust CLIP through prompt-learning \cite{menghini2023enhancing,wang2024vilt,miyai2024locoop}, fine-tuning \cite{zhang2022tip,wu2023feature,liu2024fully}, and other techniques \cite{jiao2024ai,liang2024recognizing} to enhance its generalization ability in FSL tasks \cite{ahmed2024clifs}. For example, Tip-Adapter \cite{zhang2022tip} introduces an adapter that enables CLIP to quickly adapt to new tasks and achieve good performance in FSL scenarios. However, although these methods improve the adaptability of CLIP-based FSL, the inter-class confusion problem in new tasks still exists.

\section{Our Approach}
\label{section_approach}

\begin{figure*}
    \centering
    \includegraphics[width=0.98\linewidth]{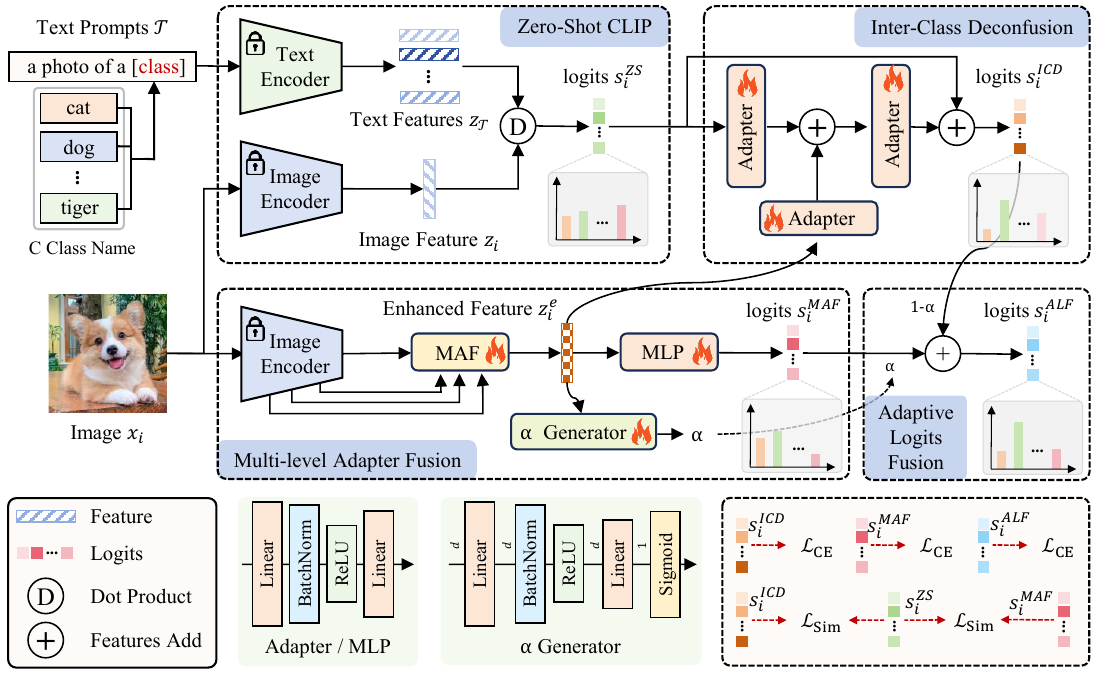}
    \caption{Overall architecture of our LDC. Our method consists of four main modules, namely Zero-Shot CLIP (ZS-CLIP), Inter-Class Deconfusion (ICD), Multi-level Adapter Fusion (MAF), and Adaptive Logits Fusion (ALF). In addition, our method includes three cross-entropy losses and two similarity losses for optimizing the learnable parameters. In ALF, the $\alpha$ Generator generates an adaptive weight $\alpha$ used to fuse the logits $s_i^{MAF}$ and $s_i^{ICD}$. All learnable parameters are in ICD, MAF, and $\alpha$ Generator. MAF is detailed in Section \ref{sec_maf}.}
    \label{fig:arch}
\end{figure*}

Our method aims to explore the transferability of CLIP with few samples so that it can be applied to new domains. 

\subsection{Problem Statement}
\label{sec_ps}

In our work, we focus on using CLIP for downstream image classification tasks in the FSL setting \cite{radford2021learning,tang2024amu}. 
Given a training set $\mathcal{D}_{train}=\{ (x_i, y_i) \}_{i=1}^{C\times K}$, where $x_i$ is an image and $y_i\in \{1,2,…,C\}$ is the label of the image $x_i$. In the training set, there are $K$ images for each class, so there are a total of $C\times K$ images for training. Our work can be simply viewed as an efficient parameter fine-tuning CLIP using $C\times K$ training images, with the goal of learning learnable parameters $\theta$ by minimizing the loss function $\mathcal{L}$:
\begin{equation}
    \begin{split}
    \mathcal{L}(\theta) = -\frac{1}{C\times K}\sum_{i=1}^K \sum_{\Bar{y}=1}^C  \log P(\Bar{y}|x_i; \theta),
    \end{split}
    \label{eq_3_1}
\end{equation}
where $P(\Bar{y}|x_i; \theta)$ is the probability that the model predicts that the image $x_i$ belongs to the class $\Bar{y}$.
To evaluate the performance of the model, a test set $\mathcal{D}_{test}=\{ (x_j, y_j) \}_{j=1}^{N}$ is given, where $x_j$ is an image and $y_j\in \{1,2,…,C\}$ is the label of the image $x_j$, $N$ is the total number of test images. For each image $x_j$ in the test set, we use the fine-tuned model to make classification predictions:
\begin{equation}
    \begin{split}
    \hat{y}_j = \arg\max_{\Bar{y}\in \{1,2,…,C\}} P(\Bar{y}|x_j; \theta),
    \end{split}
    \label{eq_3_2}
\end{equation}
and calculate the accuracy $ACC$ on the test set:
\begin{equation}
    \begin{split}
    ACC = \frac{1}{N}\sum_{j=1}^N \mathbb{I}(\hat{y}_j=y_j),
    \end{split}
    \label{eq_3_3}
\end{equation}
where $\mathbb{I}(\cdot)$ is an indicator function, which takes the value of 1 when the prediction $\hat{y}_j$ is equal to the label $y_j$ and 0 otherwise. 
Note that in the setting of this work, the test set contains the same $C$ categories as the training set.

\subsection{CLIP and Modeling Logits DeConfusion}
\label{sec_clip_mld}

In this work, our model is built on CLIP, which achieves ZSL by establishing a shared embedding space between images and text through contrastive learning \cite{radford2021learning}. Given a test image $x_i$ and a set of text prompts $\mathcal{T}=\{t_1, t_2, ..., t_C\}$ representing $C$ categories, CLIP predicts the zero-shot logits $s_i^{ZS}$ and the category $\hat{y}_i$ of the image $x_i$. Specifically, CLIP first calculates the cosine similarity $s(x_i, t_{\Bar{y}_i})$ of each image-text pair ($x_i$, $t_{\Bar{y}_i}$), then normalizes it using softmax to get the zero-shot logit $s_i^{ZS}[\Bar{y}_i]$, and finally predicts the category $\hat{y}_i$ corresponding to the maximum zero-shot logit:
\begin{equation}
    \begin{split}
    & \hat{y}_i = \arg\max_{\Bar{y}_i \in \{1,2,…,C\}} s_i^{ZS}[\Bar{y}_i], \\
    & s_i^{ZS}[\Bar{y}_i] = \frac{\exp(s(x_i, t_{\Bar{y}_i}))}{\sum_{\Bar{y}_i^{'} \in \{1,2,…,C\}} \exp (s(x_i, t_{\Bar{y}_i^{'}}))}  \\
    & s(x_i, t_{\Bar{y}_i}) = \frac{z_{x_i} \cdot z_{t_{\Bar{y}_i}}}{||z_{x_i}||||z_{t_{\Bar{y}_i}}||} \\
    \end{split}
    \label{eq_3_3_0}
\end{equation}
where $z_{x_i}=\mathcal{E}_{I}(x_i)$ and $z_{t_{\Bar{y}_i}}=\mathcal{E}_{T}(t_{\Bar{y}_i})$ are the embeddings of the image and text respectively, and $\cdot$ represents the cosine similarity. Through this process, CLIP is able to make predictions for new categories without additional training by simply adding their text prompts to $\mathcal{T}$ \cite{zhou2022learning}. 

Although CLIP has strong zero-shot capabilities, it is difficult to capture fine-grained features related to downstream tasks due to the pre-training method and data distribution, resulting in serious inter-class confusion in the predicted zero-shot logits $s_i^{ZS}$. In the experiment, we also found that there are some fixed inter-class confusion patterns for each category in $s_i^{ZS}$. To alleviate this phenomenon, we directly model and learn these patterns in logits and hope to remove them with the idea of residual learning. Specifically, we first assume that the inter-class confusion in logits can be represented as an additional noise term $\Delta s$:
\begin{equation}
    \begin{split}
    \hat{s}_i = s_i^{ZS} - \Delta s (x_i),
    \end{split}
    \label{eq_3_3_1}
\end{equation}
where $\hat{s}_i$ is the clean logits and $\Delta s (x_i)$ model the inter-class confusion of the image $x_i$. Then, we learn the additional noise term $\Delta s (x_i)$ through a learnable module $\mathcal{E}_{\Delta}$:
\begin{equation}
    \begin{split}
    \Delta s (x_i) = \mathcal{E}_{\Delta} (s_i^{ZS}, x_i).
    \end{split}
    \label{eq_3_3_2}
\end{equation}
The learnable module $\mathcal{E}_{\Delta}$ takes the image $x_i$ as prior and learns the inter-class confusion pattern from zero-shot logits $s_i^{ZS}$. 
In this way, the learnable parameters in our method can be optimized by minimizing the cross entropy loss between the clean logits $\hat{s}_i$ and the label $y_i$:
\begin{equation}
    \begin{split}
    \mathcal{L}_{CE}(\hat{s}_i, y_i) = - \log \hat{s}_i[y_i]. 
    \end{split}
    \label{eq_3_3_3}
\end{equation}
To prevent over-deconfusion, we adopt a similarity loss $\mathcal{L}_{sim}$ with $L_1$ regularization to ensure that the clean logits $\hat{s}_i$ and the original zero-shot logits $s_i^{ZS}$ maintain similar:
\begin{equation}
    \begin{split}
    \mathcal{L}_{sim}(\hat{s}_i, s_i^{ZS}) = ||\hat{s}_i - s_i^{ZS}||_1. 
    \end{split}
    \label{eq_3_3_4}
\end{equation}
Finally, the total loss for modeling Logits DeConfusion is:
\begin{equation}
    \begin{split}
    \mathcal{L} = \mathcal{L}_{CE}(\hat{s}_i, y_i) + \lambda \mathcal{L}_{sim}(\hat{s}_i, s_i^{ZS}), 
    \end{split}
    \label{eq_3_3_5}
\end{equation}
where $\lambda$ is a trade-off parameter used to balance the impact between the cross entropy loss and the similarity loss.

\subsection{Method Overview}
\label{sec_mo}

The overall architecture of our method is shown in Figure \ref{fig:arch}, which consists of four main modules: Zero-Shot CLIP (ZS-CLIP), Multi-level Adapter Fusion (MAF), Inter-Class Deconfusion (ICD), and Adaptive Logits Fusion (ALF). 

First, we obtain the zero-shot logits $s_i^{ZS}$ of the original ZS-CLIP through Eq. \ref{eq_3_3_0}. Then, through the MAF module, we transform and fuse the features of different levels of the image encoder $\mathcal{E}_I$ into an enhanced feature $z_i^e$, which integrates low-level detail information and high-level semantic information, making it more generalizable when faced with few annotated training images. After an MLP, we get the MAF logits $s_i^{MAF}$ with only the visual feature as input. Next, the ICD module uses the enhanced feature $z_i^e$ as priors and learns inter-class confusion from the zero-shot logits $s_i^{ZS}$ through a residual structure. Finally, in ALF, the MAF logits $s_i^{MAF}$ and the ICD logits $s_i^{ICD}$ are fused in a learning manner to obtain the final ALF logits $s_i^{ALF}$, where the adaptive fusion weight $\alpha$ is generated by a $\alpha$ Generator. To optimize learnable parameters in ICD, MAF, and $\alpha$ Generator, we use multiple losses to jointly learn, such as the cross entropy loss $\mathcal{L}_{CE}$ between the MAF logits $s_i^{MAF}$ and the label $y_i$, the similarity loss $\mathcal{L}_{Sim}$ between the MAF logits $s_i^{MAF}$ and the zero-shot logits $s_i^{ZS}$, etc.

\subsection{Multi-level Adapter Fusion}
\label{sec_maf}

\begin{figure}
    \centering
    \includegraphics[width=1.0\linewidth]{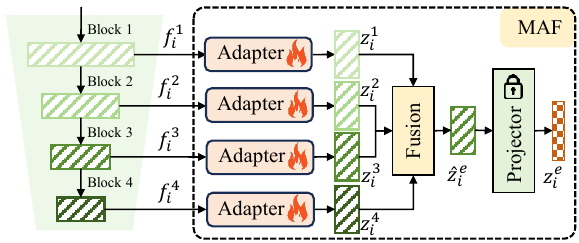}
    \caption{Details of MAF. On the left is the image encoder $\mathcal{E}_I$.}
    \label{fig:maf}
\end{figure}

Our proposed MAF module aims to fully exploit the diversity of features at each level of the image encoder $\mathcal{E}_I$. By transforming and fusing features at different levels, our MAF module not only retains the detailed information of low-level features, but also effectively utilizes the abstract semantic information of high-level features. Specifically, as shown in Figure \ref{fig:maf}, our MAF contains multiple side Adapters, a Fusion mechanism, and a Projector. First, four different levels of features $f_i^1$, $f_i^2$, $f_i^3$, and $f_i^4$ are obtained from the image encoder $\mathcal{E}_I$. Then, these four features are transformed through different Adapters to obtain four new features $z_i^1$, $z_i^2$, $z_i^3$, and $z_i^4$. Next, these features are fused into one fused feature $\hat{z}_i^e$ through our Fusion mechanism. Finally, an enhanced feature $z_i^e$ is obtained through the frozen Projector. It is worth noting that when the image encoder is based on the ResNet architecture, the Projector is an attention pooling layer, and when the image encoder is based on ViT, the Projector is a linear projection layer. 

\begin{figure}
    \centering
    \includegraphics[width=0.96\linewidth]{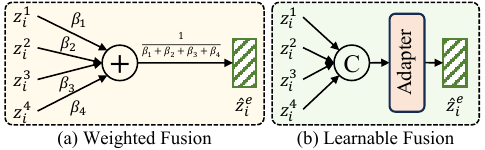}
    \caption{Details of our Fusion mechanisms.}
    \label{fig:fusion}
\end{figure}

In our MAF, we propose two different Fusion mechanisms: Weighted Fusion (WF) and Learnable Fusion (LF). As shown in Figure \ref{fig:fusion} (a), WF first fuses features of different levels through preset weights $\beta_1$, $\beta_2$, $\beta_3$, and $\beta_4$, and then get the fused feature $\hat{z}_i^e$ by dividing $\beta_1 + \beta_2 + \beta_3 + \beta_4$. As shown in Figure \ref{fig:fusion} (b), LF first concatenates features across the feature channel dimension and then uses an Adapter to reduce the dimension to obtain the fused feature $\hat{z}_i^e$, where the detailed structure of the Adapter is shown in Figure \ref{fig:arch}.

\subsection{Inter-Class Deconfusion}
\label{sec_icd}
In order to solve the inter-class confusion phenomenon that occurs when CLIP is applied to downstream tasks, we propose an Inter-Class Deconfusion (ICD) module that uses a residual structure to learn and alleviate inter-class confusion in the zero-shot logits $s_i^{ZS}$. Specifically, first, an Adapter $\mathcal{E}_{A_1}^{ICD}$ learns the inter-class confusion pattern from the zero-shot logits $s_i^{ZS}$, then another Adapter $\mathcal{E}_{A_2}^{ICD}$ learns the prior of inter-class confusion from the enhanced feature $z_i^e$, then the outputs of $\mathcal{E}_{A_1}^{ICD}$ and $\mathcal{E}_{A_2}^{ICD}$ are sent to the third Adapter $\mathcal{E}_{A_3}^{ICD}$ to jointly learn the inter-class confusion pattern from the zero-shot logits $s_i^{ZS}$ and the enhanced feature $z_i^e$, and finally the residual structure is used to remove the learned inter-class confusion pattern to obtain a relatively clean ICD logits $s_i^{ICD}$. The above process can be expressed as:
\begin{equation}
    \small
    \begin{split}
    s_i^{ICD} & = s_i^{ZS} - \mathcal{E}_{\Delta} (s_i^{ZS}, z_i^e) \\
              & = s_i^{ZS} + \mathcal{E}_{A_3}^{ICD}\big(\mathcal{E}_{A_1}^{ICD}(s_i^{ZS}) + \mathcal{E}_{A_2}^{ICD}(z_i^e)\big). 
    \end{split}
    \label{eq_3_6}
\end{equation}

Our ICD module not only takes into account the confusion cues of the zero-shot logits $s_i^{ZS}$ itself, but also introduces the enhanced feature $z_i^e$ to give cues of inter-class confusion. In this way, our ICD module learns the inter-class confusion patterns and weakens such patterns in the form of residuals, making better use of the knowledge learned from CLIP and improving the modeling ability of new domain data, especially in the FSL setting.

\subsection{Adaptive Logits Fusion}
\label{sec_alf}

After obtaining the ICD logits $s_i^{ICD}$ from the ICD module, we further propose an Adaptive Logits Fusion (ALF) module, which fuses the ICD logits $s_i^{ICD}$ with the MAF logits $s_i^{MAF}$ in an adaptive manner to obtain the final ALF logits $s_i^{ALF}$. Specifically, we first design a $\alpha$ Generator that generates an adaptive weight $\alpha$ using the enhanced feature $z_i^e$, and then use the adaptive weight $\alpha$ to weightedly fuse the ICD logits $s_i^{ICD}$ and the MAF logits $s_i^{MAF}$:
\begin{equation}
    \begin{split}
    s_i^{ALF} = \alpha s_i^{MAF} + (1 - \alpha) s_i^{ICD}, \ \ \alpha = \mathcal{E}_{G_\alpha}(z_i^e),  
    \end{split}
    \label{eq_3_7}
\end{equation}
where $ \mathcal{E}_{G_\alpha}$ is the $\alpha$ Generator, as shown in Figure \ref{fig:arch}. 

Through the above ALF module, our method can adaptively combine inter-class confusion with multi-level features, making the final ALF logits $s_i^{ALF}$ more robust and accurate. The core of this is to use the adaptive weight $\alpha$ to adaptively adjust the weight between the ICD logits and the MAF logits to ensure a more reasonable fusion.

\subsection{Training Objective}
\label{sec_to}

In the optimization stage, we use the cross-entropy loss ($\mathcal{L}_{CE}$) and the $L_1$ similarity loss ($\mathcal{L}_{Sim}$) to jointly optimize the learnable parameters of our method. Specifically, we  use Eq. \ref{eq_3_3_3} to calculate the loss $\mathcal{L}_{CE}^{MAF}$ between $s_i^{MAF}$ and $y_i$, the loss $\mathcal{L}_{CE}^{ICD}$ between $s_i^{ICD}$ and $y_i$, and the loss $\mathcal{L}_{CE}^{ALF}$ between $s_i^{ALF}$ and $y_i$ to ensure the classification accuracy: 
\begin{equation}
    \begin{split}
    \mathcal{L}_{CE} = & \mathcal{L}_{CE}^{MAF}(s_i^{MAF}, y_i) + \\ 
                       & \mathcal{L}_{CE}^{ICD}(s_i^{ICD}, y_i) + \mathcal{L}_{CE}^{ALF}(s_i^{ALF}, y_i). 
    \end{split}
    \label{eq_3_8}
\end{equation}
In addition, in order to avoid over-deconfusion, we keep the similarity between the output logits and the original zero-shot logits. Specifically, we use Eq. \ref{eq_3_3_4} to calculate the similarity loss $\mathcal{L}_{Sim}^{MAF}$ between $s_i^{MAF}$ and $s_i^{ZS}$ and the similarity loss $\mathcal{L}_{Sim}^{ICD}$ between $s_i^{ICD}$ and $s_i^{ZS}$:
\begin{equation}
    \begin{split}
    \mathcal{L}_{Sim} = \mathcal{L}_{Sim}^{MAF} + \mathcal{L}_{Sim}^{ICD}. 
    \end{split}
    \label{eq_3_9}
\end{equation}
Finally, as shown in Eq. \ref{eq_3_3_5}, the total loss $\mathcal{L}$ is:
\begin{equation}
    \begin{split}
    \mathcal{L} = \mathcal{L}_{CE} + \lambda \mathcal{L}_{Sim}. 
    \end{split}
    \label{eq_3_10}
\end{equation}
where $\lambda$ is a trade-off parameter.

\section{Experimental Results}
\label{section_result}

In this section, we conduct extensive experiments to verify the effectiveness of the proposed method and illustrate the contribution of each module to the performance.

\subsection{Datasets}

To validate our method, we consider 11 image classification benchmarks: ImageNet \cite{deng2009imagenet}, Caltech101 \cite{fei2004learning}, DTD \cite{cimpoi2014describing}, EuroSAT \cite{helber2019eurosat}, FGVCAircraft \cite{maji2013fine}, Flowers102 \cite{nilsback2008automated}, Food101 \cite{bossard2014food}, OxfordPets \cite{parkhi2012cats}, StanfordCars \cite{krause20133d}, SUN397 \cite{xiao2010sun}, and UCF101 \cite{soomro2012ucf101}. In addition, in order to verify the generalization ability, we use ImageNet \cite{deng2009imagenet} as training data and ImageNet-Sketch \cite{wang2019learning} and ImageNet-V2 \cite{recht2019imagenet} benchmarks as out-of-domain distributions (OOD). We follow the FSL setting of CLIP \cite{radford2021learning} and use 1, 2, 4, 8, and 16 images per class to train our model and evaluate the results on the test set. Following the data splitting strategy of previous methods, such as CoOp \cite{zhou2022learning} and Tip-Adapter \cite{zhang2022tip}, we divide each dataset into training, validation, and test sets.

\subsection{Implementation Details}

For fair comparison, our experiments are performed on ResNet-50 \cite{he2016deep} unless otherwise specified. We train all tasks for 50 epochs and use AdamW \cite{parikh2014proximal} with an initial learning rate of 0.001 and a batch size of 64. The input size of all datasets is 224, and random resized crop and horizontal flip are used as augmentation methods. In WF, $\beta_1$, $\beta_2$, $\beta_3$, and $\beta_4$ are set to 0.1, 0.2, 0.3, and 0.4, respectively. In the loss, the value of $\lambda$ is 1.0. All experiments are implemented based on PyTorch \cite{paszke2019pytorch} and performed on a single NVIDIA GTX 4090D GPU. The training and testing time on all datasets for the 16-shot setting is about 37 minutes.

\subsection{Comparison with the State-of-the-art}

The effectiveness of our method is evaluated in two settings: downstream classification benchmarks and OOD data.

\begin{figure*}
    \centering
    \includegraphics[width=1.0\linewidth]{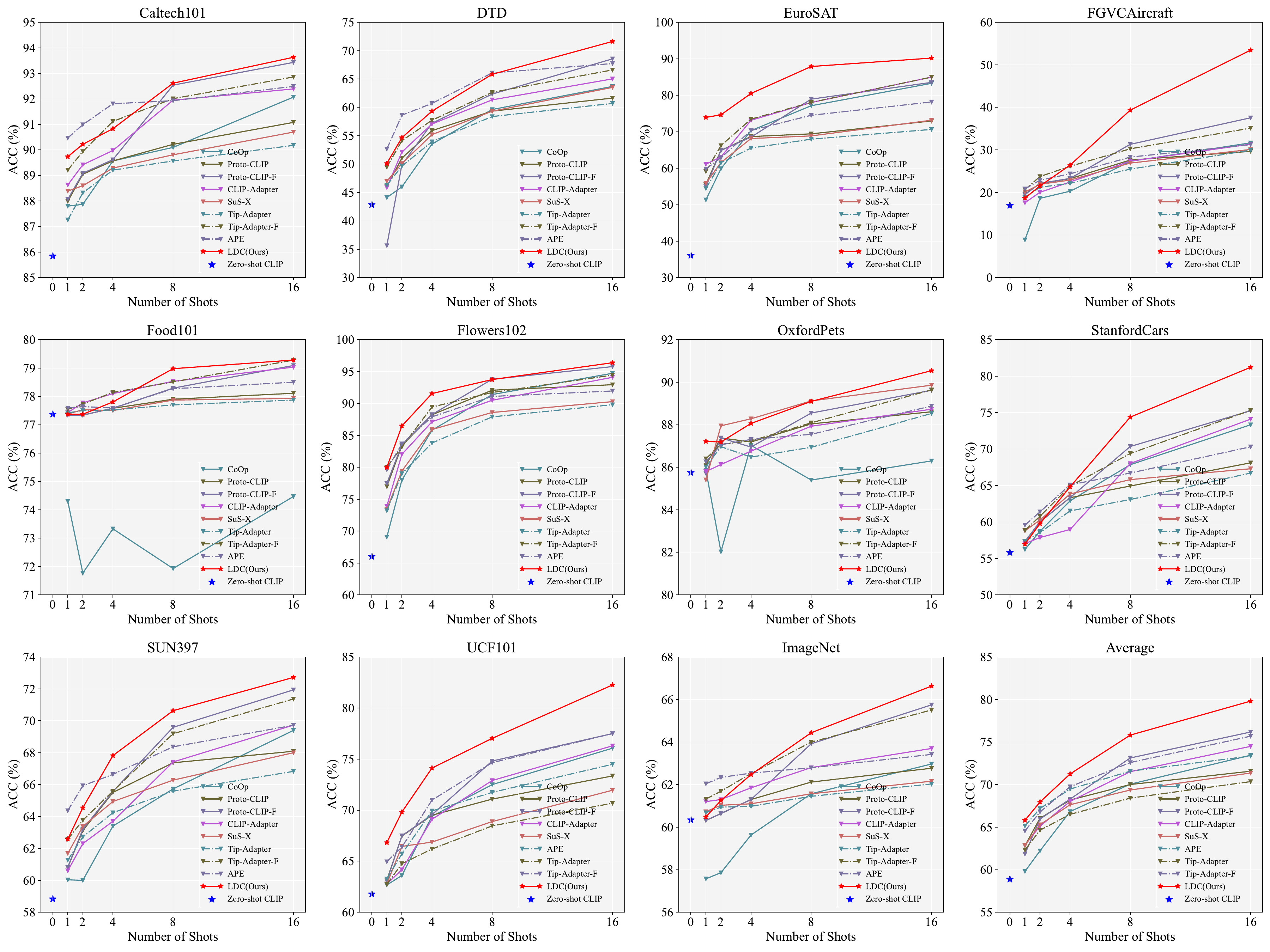}
    \caption{Classification performance of different methods on 11 datasets, and the last one is the average performance on these 11 datasets.}
    \label{fig:result}
\end{figure*}

\begin{table}[t]
    \centering
 \resizebox{0.95\linewidth}{!}{
    \begin{tabular}{lccccc}
        \toprule
        \multirow{2}{*}{Method} & \multicolumn{5}{c}{K-Shot Accuracy (\%)} \\
         & 1 & 2 & 4 & 8 & 16 \\
        \midrule
        LP-CLIP \cite{radford2021learning} & 22.17 & 31.90 & 41.20 & 49.52 & 56.13 \\
        CoOp \cite{zhou2022learning} & 57.15 & 57.81 & 59.99 & 61.56 & 62.95 \\
        VT-CLIP \cite{qiu2021vt} & 60.53 & 61.29 & 62.02 & 62.81 & 63.92 \\
        Tip-Adapter \cite{zhang2022tip} & 60.70 & 60.92 & 60.95 & 61.48 & 62.00 \\
        Tip-Adapter-F \cite{zhang2022tip} & 61.32 & 61.69 & 62.52 & 64.00 & 65.51 \\
        SuS-X \cite{udandarao2023sus} & 60.73 & 61.03 & 61.10 & 61.57 & 62.16 \\
        FAR \cite{wu2023feature} & 60.80 & 61.90 & 62.40 & \underline{64.30} & \underline{66.39}  \\
        CALIP-FS \cite{guo2023calip} & \underline{61.35} & 62.03 & \textbf{63.13} & 64.11 & 65.81 \\
        SGVA-CLIP \cite{peng2023sgva} & 61.00 & \textbf{62.70} & \underline{63.10} & 64.20 & 65.70 \\
        Proto-CLIP-F \cite{palanisamy2023proto} & 60.32 & 60.64 & 61.30 & 63.92 & 65.75 \\
        APE \cite{zhu2023not} & \textbf{62.04} & \underline{62.34} & 62.54 & 62.79 & 63.42 \\
        CLIP-Adapter \cite{gao2024clip} & 61.20 & 61.52 & 61.84 & 62.68 & 63.59 \\
        DAC-V \cite{gondal2024domain} & 60.71 & 61.48 & 61.87 & 63.38 & 64.89 \\
        LP++ \cite{huang2024lp++} & 61.20 & 61.60 & 62.60 & 63.80 & 64.70 \\
        \midrule
        \multirow{2}{*}{LDC (Our)} & 60.48 & 61.25 & 62.47 & \textbf{64.44} & \textbf{66.63} \\
         & \alisurebed{-1.56} & \alisurebed{-1.45} & \alisurebed{-0.66} & \alisuregood{+0.14} & \alisuregood{+0.24} \\
        \bottomrule
    \end{tabular}
}
    \caption{Comparison of different methods on ImageNet.}
    \label{table0}
\end{table}

\begin{table}[t]
    \centering
 \resizebox{1.0\linewidth}{!}{
    \begin{tabular}{lcccccc}
        \toprule
        \multirow{2}{*}{Method} & \multicolumn{6}{c}{K-Shot Accuracy (\%)} \\
        & 0 & 1 & 2 & 4 & 8 & 16 \\
        \midrule
        ZS-CLIP & 58.87 & - & - & - & - & - \\
        \midrule
        LP-CLIP \cite{radford2021learning} & - & 36.67 & 47.61 & 57.19 & 64.98 & 71.10 \\
        CoOp \cite{zhou2022learning} & - & 59.80 & 62.21 & 66.84 & 70.05 & 73.45 \\
        Tip-Adapter \cite{zhang2022tip} & - & 62.32 & 64.64 & 66.51 & 68.41 & 70.36 \\
        Tip-Adapter-F \cite{zhang2022tip} & - & 64.55 & 66.79 & \underline{69.76} & 72.59 & 75.69 \\
        SuS-X \cite{udandarao2023sus} & - & 62.87 & 65.29 & 67.64 & 69.37 & 71.36 \\
        APE \cite{zhu2023not} & - & \underline{65.13} & \underline{67.19} & 69.47 & 71.58 & 73.36 \\
        CLIP-Adapter \cite{gao2024clip} & - & 62.90 & 65.11 & 68.02 & 71.52 & 74.50 \\
        Proto-CLIP \cite{palanisamy2023proto} & - & 62.83 & 66.04 & 68.19 & 70.01 & 71.57 \\
        Proto-CLIP-F \cite{palanisamy2023proto} & - & 61.84 & 65.96 & 68.29 & \underline{73.13} & \underline{76.18} \\
        \midrule
        \multirow{2}{*}{LDC (Our)} &  - &\textbf{65.71} & \textbf{67.92} & \textbf{71.17} & \textbf{75.79} & \textbf{79.78} \\
         & - & \alisuregood{+0.58} & \alisuregood{+0.73} & \alisuregood{+1.41} & \alisuregood{+2.66} & \alisuregood{+3.60} \\
        \bottomrule
    \end{tabular}
}
    \caption{Comparison of different methods over 11 datasets.}
    \label{table1}
\end{table}

\begin{table}[t]
    \centering
 \resizebox{0.90\linewidth}{!}{
    \begin{tabular}{clccc}
        \toprule
         & \multirow{2}{*}{Method} & Source & \multicolumn{2}{c}{Target} \\
        \cline{3-5}
         & & ImageNet & V2 & Sketch \\
        \midrule
        \multirow{14}{*}{\rotatebox{90}{ResNet-50}}
        & ZS-CLIP \cite{radford2021learning} & 60.33 & 53.27 & 35.44 \\
        & LP-CLIP \cite{radford2021learning} & 56.13 & 45.61 & 19.13 \\
        & CoOp \cite{zhou2022learning} & 62.95 & 54.58 & 31.04 \\
        & CoCoOp \cite{zhou2022conditional} & 62.81 & 55.72 & 34.48 \\
        & CALIP-FS \cite{guo2023calip} & 65.81 & 55.98 & 35.37 \\
        & CLIP-Adapter \cite{gao2024clip} & 63.59 & 55.69 & 35.68 \\
        & APE \cite{zhu2023not} & 63.42 & 55.94 & \underline{36.61} \\
        & APE-T \cite{zhu2023not} & 66.07 & 57.59 & 36.36 \\
        & Tip-Adapter \cite{zhang2022tip} & 62.03 & 54.60 & 35.90 \\
        & Tip-Adapter-F \cite{zhang2022tip} & 65.51 & 57.11 & 36.00 \\
        & DAC-V \cite{gondal2024domain} & 64.89 & 56.56 & 36.27  \\
        & DAC-VT \cite{gondal2024domain} & \underline{66.61} & 57.68 & 35.33  \\
        & FAR \cite{wu2023feature} & 66.39 & \underline{57.91} & \textbf{36.69} \\
        \cline{2-5}
        \noalign{\vspace{1mm}}
        & LDC (Our) & \textbf{66.63} \alisuregood{\scriptsize{+0.02}} & \textbf{58.03} \alisuregood{\scriptsize{+0.12}} & 35.52 \alisurebed{\scriptsize{-1.17}} \\
        \midrule
        \multirow{6}{*}{\rotatebox{90}{ViT-B/16}}
        & ZS-CLIP \cite{radford2021learning} & 66.73 & 60.83 & 46.15 \\
        & CoOp \cite{zhou2022learning} & 71.51 & 64.20 & 47.99 \\
        & CoCoOp \cite{zhou2022conditional} & 71.02 & 64.07 & 48.75 \\
        & MaPLe \cite{khattak2023maple} & 70.72 & 64.07 & \underline{49.15} \\
        & RPO \cite{lee2023read} & \underline{71.67} & \underline{65.13} & \textbf{49.27} \\
        \cline{2-5}
        \noalign{\vspace{1mm}}
        & LDC (Our) & \textbf{73.88} \alisuregood{\scriptsize{+2.21}} & \textbf{66.10} \alisuregood{\scriptsize{+0.97}} & 48.85 \alisurebed{\scriptsize{-0.42}} \\
        \bottomrule
    \end{tabular}
}
    \caption{Comparison of different methods under OOD setting.}
    \label{table2}
\end{table}

\subsubsection{Results on Classification Benchmarks} 

To verify the effectiveness of our LDC, we compare it with several SOTA CLIP-based FSL methods with ResNet-50 \cite{he2016deep} as the backbone on ImageNet \cite{deng2009imagenet}, such as CoOp \cite{zhou2022learning}, VT-CLIP \cite{qiu2021vt}, Tip-Adapter \cite{zhang2022tip}, SuS-X \cite{udandarao2023sus}, FAR \cite{wu2023feature}, CALIP-FS \cite{guo2023calip}, SGVA-CLIP \cite{peng2023sgva}, Proto-CLIP-F \cite{palanisamy2023proto}, APE \cite{zhu2023not}, DAC-V \cite{gondal2024domain}, and LP++ \cite{huang2024lp++}. Experimental results are shown in Table \ref{table0}. It can be seen that our method has achieved the best accuracy in some cases. For example, in the 16-shot setting, we achieve an accuracy of 66.63\%, which is 0.24 higher than the previous best method. To comprehensively evaluate our method on a variety of downstream tasks, we summarize the average FSL classification performance of various methods with 1, 2, 4, 8, and 16-shot settings on 11 datasets in Table \ref{table1}. Overall, our method is the best on all settings, especially on the 16-shot setting, where we achieve an overall accuracy of 79.78\%. These results show that our LDC can effectively model and learn inter-class confusion patterns and remove them. 
In addition, Figure \ref{fig:result} shows the detailed accuracy of our method and various comparison methods on 11 datasets, where the last one shows the average performance on these 11 datasets. It can be seen that the results of our method are excellent in most cases, especially on the EuroSAT \cite{helber2019eurosat}, FGVCAircraft \cite{maji2013fine}, StanfordCars \cite{krause20133d}, and UCF101 \cite{soomro2012ucf101} datasets.

\subsubsection{Robustness to OOD Data} 

To verify the robustness of our method, we first train our model on ImageNet  \cite{deng2009imagenet} in a 16-shot setting, and then test it on ImageNet-V2 (V2) \cite{recht2019imagenet} and ImageNet-Sketch (Sketch) \cite{wang2019learning} to evaluate the performance of our method under OOD setting. Table \ref{table2} shows the performance of our method and other methods under two backbones, ResNet-50 \cite{he2016deep} and ViT-B/16 \cite{dosovitskiy2020image}. It can be seen that our method performs well on ImageNet-V2 with both backbones, which confirms the robustness and transferability of our LDC. However, the performance on ImageNet-Sketch is not as expected, which indicates that our method may not be well suited for tasks with large domain differences and needs further study.

\subsection{Ablation Experiment}

To analyze the effect of our modules and settings, we conduct a series of ablation experiments over 11 datasets.

\subsubsection{Effect analysis of each module} 
\label{ae_1}

In our work, we mainly propose three modules, MAF, ICD, and ALF. To explore their respective roles, we perform a set of module ablation experiments under multiple backbones. The experimental results are shown in Table \ref{abl_1}. As can be seen from the table, our ICD significantly improves the average accuracy by 19.07\% with ResNet50 \cite{he2016deep} as the backbone and 16.46\% with ViT-B/16 \cite{dosovitskiy2020image} as the backbone, which shows that our method can learn the inter-class confusion patterns and remove them, thus achieving our motivation.

\begin{table}[t]
\centering
 \resizebox{0.78\linewidth}{!}{
	\begin{tabular}{ccc|ll}
		\hline
        \multicolumn{3}{c|}{Module} & \multicolumn{2}{c}{Backbone} \\
		MAF & ICD & ALF & ResNet50 & ViT-B/16 \\
		\hline
         & & & 58.87 & 65.52 \\
        \checkmark & & & 70.83 \alisuregood{\scriptsize{+11.96}} & 74.57 \alisuregood{\scriptsize{+9.05}} \\
        & \checkmark & & 77.94 \alisuregood{\scriptsize{+19.07}} & 81.98 \alisuregood{\scriptsize{+16.46}} \\
        \checkmark & \checkmark & & 78.82 \alisuregood{\scriptsize{+19.95}} & 82.41 \alisuregood{\scriptsize{+16.89}} \\
        \checkmark & \checkmark & \checkmark & 79.78 \alisuregood{\scriptsize{+20.91}} & 82.87 \alisuregood{\scriptsize{+17.35}} \\
		\hline
	\end{tabular}
 }
\caption{Effect analysis of each module over 11 datasets.} 
\label{abl_1}
\end{table}

\subsubsection{Ablation study of MAF}

The ablation results of our MAF module are shown in Table \ref{abl_2}, including the ablation of multi-level features, fusion mechanisms, and projector. From the left of Table \ref{abl_2}, it can be seen that as the number of feature levels increases, the average performance improves, which is consistent with intuition and previous experience \cite{lin2017feature,li2022mfnet}. In the left of Table \ref{abl_2}, "WF (0.25)" indicates $\beta_1$=$\beta_2$=$\beta_3$=$\beta_4$=0.25, and "WF (0.1/0.2/0.3/0.4)" indicates $\beta_1$=0.1, $\beta_2$=0.2, $\beta_3$=0.3, and $\beta_4$=0.4. From the results, we can observe that among our proposed fusion mechanisms, "WF (0.1/0.2/0.3/0.4)" achieves the best results. In addition, after fusion, the projector also plays an important role in the performance.

\begin{table}[t]
    \centering
    \begin{minipage}[b]{0.455\linewidth}  
        \centering
        \resizebox{1.0\linewidth}{!}{
        	\begin{tabular}{cccc|c}
        		\hline
        		$f_1$ & $f_2$ & $f_3$ & $f_4$ & ACC \\
        		\hline
        	  & & & \checkmark & 75.95 \\
        	  & & \checkmark & \checkmark & 77.48 \\
        	  & \checkmark & \checkmark & \checkmark & 79.10 \\
        	  \checkmark & \checkmark & \checkmark & \checkmark & 79.78 \\
        		\hline
        	\end{tabular}
        }
    \end{minipage}
    \hfill
    \begin{minipage}[b]{0.535\linewidth}  
        \centering
        \resizebox{1.0\linewidth}{!}{
            \begin{tabular}{l|c}
                \hline
                w/ or w/o & ACC \\
                \hline
                w/ WF (0.25)& 79.71 \\
                w/ WF (0.1/0.2/0.3/0.4) & 79.78 \\
                w/ LF & 79.59 \\
                w/o Projector & 78.94 \\
                \hline
            \end{tabular}
        }
    \end{minipage}
    \caption{Ablation study of MAF over 11 datasets.} 
    \label{abl_2}
\end{table}

\subsubsection{Ablation study of ICD}

The ablation experimental results of ICD are shown in Table \ref{abl_3_1}, where "$A_1$" means $\mathcal{E}_{A_1}^{ICD}(s_i^{ZS})$, "$A_2$" means $\mathcal{E}_{A_2}^{ICD}(z_i^e)$, "$A_3$" means $\mathcal{E}_{A_3}^{ICD}(\cdot)$, and "Res" means the residual structure. It can be seen that using the enhanced feature $z_i^e$ as prior can improve the average performance. In addition, it can be seen from the last two rows that the residual structure can improve the performance by 1.24\%.

\begin{table}[t]
    \centering
    \begin{minipage}[b]{0.473\linewidth}  
        \centering
        \resizebox{1.0\linewidth}{!}{
        	\begin{tabular}{ccc|c|c}
        		\hline
        		$A_1$ & $A_2$ & $A_3$ & \small{Res} & ACC \\
        		\hline
            	\checkmark & & \checkmark & \checkmark & 78.02 \\
            	& \checkmark & \checkmark & \checkmark & 79.36 \\
            	\checkmark & \checkmark & & \checkmark & 79.12 \\
            	\checkmark & \checkmark & \checkmark & & 78.54 \\
            	\checkmark & \checkmark & \checkmark & \checkmark & 79.78 \\
        		\hline
        	\end{tabular}
        }
        \caption{ICD ablation.} 
        \label{abl_3_1}
    \end{minipage}
    \hfill
    \begin{minipage}[b]{0.515\linewidth}  
        \centering
        \resizebox{1.0\linewidth}{!}{
        	\begin{tabular}{l|c}
        		\hline
        		ALF logits & ACC \\
        		\hline
        		$s_i^{ICD}$ & 78.71 \\
        		$s_i^{MAF}$ & 77.35 \\
        		$s_i^{ICD} + s_i^{MAF}$ & 78.69 \\
        		$(s_i^{ICD} + s_i^{MAF})/2$ & 79.24 \\
        		$\alpha s_i^{ICD} + (1 - \alpha ) s_i^{MAF}$ & 79.78 \\
        		\hline
        	\end{tabular}
        }
        \caption{ALF logits ablation.} 
        \label{abl_3_3}
    \end{minipage}
    \label{abl_3}
\end{table}

\subsubsection{Ablation study of ALF}

The ablation experimental results of ALF are shown in Table \ref{abl_3_3}, where $\alpha$ is generated by the $\alpha$ Generator. Compared to manually setting $\alpha$ to 0.5 in the 4-th row, our adaptive logits fusion improves by 0.54\%. It also can be seen that fusing $s_i^{ICD}$ and $s_i^{MAF}$ can improve the average accuracy, regardless of whether the fusion weight is fixed or adaptive.

\subsubsection{Ablation study of Training Losses}
Our method contains multiple losses: three cross entropy losses for classification and two similarity losses to avoid over-deconfusion. Table \ref{abl_4} shows a set of ablation experiments on the losses. As can be seen from the table, similarity losses $\mathcal{L}_{Sim}^{MAF}$ and $\mathcal{L}_{Sim}^{ICD}$ play an important role. 

\begin{table}[t]
\vspace{-3mm}
\centering
 \resizebox{0.75\linewidth}{!}{
	\begin{tabular}{ccccc|c}
		\hline
	  $\mathcal{L}_{CE}^{MAF}$ & $\mathcal{L}_{CE}^{ICD}$ & $\mathcal{L}_{CE}^{ALF}$ & $\mathcal{L}_{Sim}^{MAF}$ & $\mathcal{L}_{Sim}^{ICD}$ & ACC \\
		\hline
        \checkmark & \checkmark & & & & 79.03 \\
        & & \checkmark & & & 75.47 \\
        \checkmark & \checkmark & \checkmark & & & 79.13 \\
		\hline
        \checkmark & \checkmark & \checkmark & \checkmark & & 79.29 \\
        \checkmark & \checkmark & \checkmark & & \checkmark & 79.43 \\
        \checkmark & \checkmark & \checkmark & \checkmark & \checkmark & 79.78 \\
		\hline
	\end{tabular}
 }
\caption{Ablation study of training losses over 11 datasets. } 
\label{abl_4}
\vspace{-2mm}
\end{table}

\section{Conclusion}
\label{section_conclusion}

This work proposes a novel Logits DeConfusion method to solve the inter-class confusion problem in CLIP-based few-shot learning tasks. Through the joint design of the Multi-level Adapter Fusion (MAF) module and the Inter-class Deconfusion (ICD) module, we effectively alleviate the inter-class confusion caused by the lack of clear classification boundaries during CLIP pre-training. The MAF module enhances the adaptability of image representation by extracting and fusing features from different levels, while the ICD module learns inter-class confusion through a learnable module with residuals, improving the clarity and accuracy of classification logits. Experimental results show that the proposed method significantly improves the classification accuracy on multiple benchmarks, verifying its effectiveness in dealing with inter-class confusion.

\noindent\textbf{Acknowledgement:} This work was supported Postdoctoral Fellowship Program of China Postdoctoral Science Foundation (CPSF) (GZC20232033), National Natural Science Foundation of China (62406231), China Postdoctoral Science Foundation (2023M742738), Shaanxi Province Postdoctoral Research Project Funding (2024BSHSDZZ119), Joint Fund Project of National Natural Science Foundation of China (U22B2054), Program for Cheung Kong Scholars and Innovative Research Team in University (IRT\_15R53), Fund for Foreign Scholars in University Research and Teaching Programs (the 111 Project) (B07048).

{\small\bibliographystyle{ieeenat_fullname}\bibliography{main_final}}
\end{document}